\documentclass{article}

\usepackage[preprint]{neurips_2026}

\usepackage[utf8]{inputenc} 
\usepackage[T1]{fontenc}    
\usepackage{hyperref}       
\usepackage{url}            
\usepackage{booktabs}       
\usepackage{amsfonts}       
\usepackage{nicefrac}       
\usepackage{microtype}      
\usepackage{xcolor}         

\usepackage{adjustbox}
\usepackage{booktabs} 
\usepackage{multirow}
\usepackage{amsmath}
\usepackage{wrapfig}
\usepackage{graphicx}
\usepackage{amssymb}
\usepackage[most]{tcolorbox}

\title{Towards Generalization of Block Attention via Automatic Segmentation and Block Distillation}

\author{%
  Shuaiyi Li$^1$\thanks{Work done during the Internship at Tencent}\\
  \And
  Zhisong Zhang$^2$\thanks{Corresponding author.} \\
  \And
  Yan Wang$^3$ \\
  \And
  Lei Zhu$^3$ \\
  \And
  Dongyang Ma$^3$ \\
  \And
  Chenlong Deng$^4$ \\
  \And
  Yang Deng$^5$ \\
  \And
  Wai Lam$^{1\dagger}$ \\
  \AND
  \qquad \qquad \texttt{\{sli, wlam\}@se.cuhk.edu.hk}, 
  \quad \texttt{zhisong.zhang@cityu.edu.hk}
  \AND
  $^1$The Chinese University of Hong Kong, \ $^2$City University of Hong Kong, \ $^3$Tencent
  \AND
  $^4$Gaoling School of Artificial Intelligence, Renmin University of China \AND
  $^5$Singapore Management University \AND
  \href{https://huggingface.co/collections/Syon-Li/generalization-of-block-attention}{\textit{HF Repo}} \quad | \quad \href{https://www.modelscope.cn/collections/SyonLi/Generalization-of-Block-Attention}{\textit{ModelScope Repo}} 
  \quad | \quad \href{https://github.com/Syon-Li/Generalization-of-Block-Attention}{\textit{Github Repo}}
}

\begin{document}
\bibliographystyle{plainnat}

\maketitle

\begin{abstract}
    Block attention, which processes the input as separate blocks that cannot attend to one another, offers significant potential to improve KV cache reuse in long-context scenarios such as Retrieval-Augmented Generation (RAG). However, its broader application is hindered by two key challenges: the difficulty of segmenting input text into meaningful, self-contained blocks, and the inefficiency of existing block fine-tuning methods that risk degrading performance. To address these, we first construct SemanticSeg, a large and diverse semantic segmentation dataset containing over $30k$ instances across 16 categories—including books, code, web text, and conversations with text lengths ranging from $2k$ to $32k$. Using this dataset, we train a lightweight segmenter to automatically partition text into human-instinct-aligned blocks with controllable granularity. Second, we propose block distillation, a training framework that is more efficient than block fine-tuning, which uses a frozen full-attention teacher model to guide the block-attention student. This framework integrates three novel components: block sink tokens to mitigate information loss at block boundaries, block dropout to leverage training signals from all blocks, and token-level loss weighting to focus learning on block-attention-sensitive tokens. Experiments across multiple models and benchmarks demonstrate that our segmenter outperforms heuristic and statistical baselines, and block distillation achieves near-full-attention performance under block attention, establishing a practical and scalable pathway for deploying block attention.
\end{abstract}

\section{Introduction}

Large language models (LLMs) have demonstrated remarkable capabilities in processing long-context inputs \citep{longbenchv2, longbench, locomo}, enabling applications such as multi-document question answering, coding, etc. However, the standard full-attention mechanism scales quadratically with sequence length, making inference on long inputs computationally expensive and memory-intensive. A significant source of this inefficiency is the context-dependent nature of full attention: when identical context is paired with different prefixes, its key-value (KV) states must be recomputed from scratch. This leads to substantial waste of compute and energy, particularly in retrieval-augmented generation (RAG) scenarios where overlapping document sets are repeatedly processed across queries. To mitigate this, block attention \citep{block-attention} has emerged as a promising alternative. By restricting self-attention to independent blocks and allowing only a final aggregation block to attend globally, it eliminates cross-block dependencies and facilitates the reuse of pre-computed KV cache. Nevertheless, its practical adoption is hindered by several obstacles.

An important barrier in the application of block attention is segmentation, that is, how to divide the input sequence into separate blocks. Existing approaches often rely on heuristic rules, such as splitting with newlines; however, such rules rarely generalize and are highly likely to break the semantic coherence of the inputs. As demonstrated in Section \ref{sec.impact-segmentation}, such naive segmentation leads to performance degradation, highlighting the need for a semantic-aware approach.
To address this, we propose a robust, data-driven semantic segmenter capable of handling diverse input formats. We take a data-driven approach and construct a segmentation dataset (\textit{SemanticSeg}), where each sample is segmented by the semantic meaning. Using this dataset, we train a neural segmenter that can automatically produce adaptive and context-aware boundaries, overcoming a major obstacle to the generalization of block attention.

Another major challenge is effectively integrating block attention into existing LLMs. While training-free strategies such as Prompt Cache \citep{promptcache} and Superposition prompting \citep{Superposition-Prompting} attempt to enable KV state reuse or parallel processing, such direct application of block attention into general domains suffers from severe performance degradation compared to full attention (Table \ref{tab:main longbench results}). This indicates that these approaches are not sufficiently effective for supporting reliable block attention, further highlighting the necessity of specialized training. \citep{block-attention} addressed this through ``block fine-tuning'', which trains models on both block and full attention patterns to balance specialized performance with general capability. However, this approach is computationally expensive and generalizes poorly across diverse domains (Section \ref{sec.weaknesses of block attention}). To address these limitations, we introduce \textit{Block Distillation}, a training framework designed for higher efficiency and signal density.
It incorporates three novel mechanisms: \emph{block sink tokens} to counteract information loss at block boundaries, \emph{block dropout} to exploit training signals from all blocks, and \emph{token-level loss weighting} to emphasize tokens that are most sensitive to block attention.

To validate our approach, we conduct comprehensive experiments across multiple models and benchmarks. To verify the effectiveness of our segmenter, we quantify the impact of segmentation on downstream performance and compare our segmenter against a range of heuristic and statistical segmentation baselines. To evaluate our training framework, we rigorously tested \textit{Block Distillation}, evaluating its efficiency and performance in general domains and the specific contributions of its individual components. Experimental results demonstrate that our segmenter consistently outperforms all baselines, and \textit{Block Distillation} pushes block-attention performance close to the full-attention upper bound while preserving or even improving full-attention capability, establishing a practical and scalable pathway for deploying block attention in long-context applications. We discuss the potential application of the block attention in Appendix \ref{appendix.application-analysis}.

\section{Preliminary}
We illustrate the idea of block attention with the example of RAG. Consider a pool of $r$ documents and two instances with overlapping retrieved contexts:
\begin{quote}
    Inst 1: Document [$i$]; Document [$i+1$]; \dots; Document [$i+x$]; \dots; [Query $q_x$]. \\
    Inst 2: Document [$j$]; Document [$j+1$]; \dots; Document [$j+y$]; \dots; [Query $q_y$].
\end{quote}
where the document sets intersect at $\{i, \dots, i+x\} \cap \{j, \dots, j+y\} = \{n, \dots, m\}$.
In standard full attention, the KV states for this intersection must be recomputed for each query because the attention mechanism is prefix-dependent, resulting in significant computational overhead and energy waste. However, if encoding can be performed independently for each document, the KV states for the intersection $\{n, \dots, m\}$ become prefix-agnostic and can be safely reused across disparate queries. Block attention \citep{block-attention} formalizes this by partitioning the input into independent blocks. Each block employs self-attention restricted to its own tokens, ensuring that internal representations are decoupled from other blocks. Only the final block (typically the user query) is permitted to utilize full attention, aggregating information from all preceding KV caches.

The implementation of this method can be achieved easily via the following steps: 1) Independently encoding each block except the last one; 2) Computing the positional encoding for each token based on their position in the input text; 3) Concatenating all pre-computed KV states of the blocks and using them to compute the KV states for the final block. However, the previous work \citep{block-attention} has demonstrated that the model cannot accommodate this pattern without training, which is also verified in this work (section \ref{sec.impact-segmentation}). To cope with this challenge, they employ block fine-tuning, which updates the parameters in both ways, one for block attention and one for full attention. They claim this could enhance block-attention performance while maintaining full attention capability. However, despite its heavy updating scheme, it struggles to generalize to other domains (\ref{sec.main-results}).

\section{Automatic Segmentation}

A major barrier to the general application of block attention is segmentation, i.e., given the input text, how to cut it into meaningful, self-contained, and human-instinct-aligned blocks (or chunks) for later processing. One may argue that the segmentation operation has a limited effect on the final performance, as the model may not understand the semantics in the same way as humans. However, as proved in the section \ref{sec.impact-segmentation}, the segmentation plays an important role in the final performance.

\begin{wrapfigure}{r}{0.6\textwidth}
\centering
\includegraphics[width=0.6\textwidth]{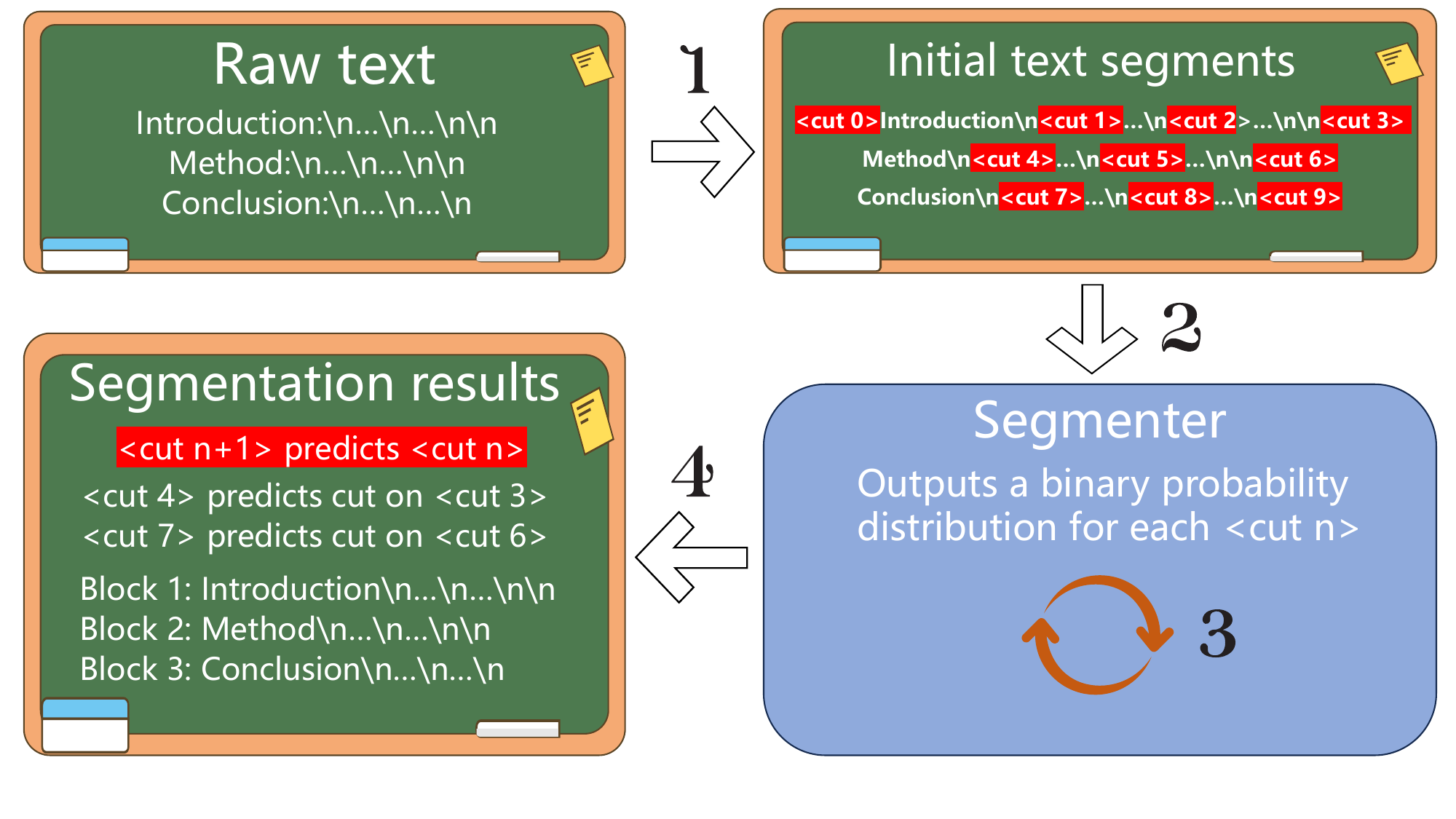}
\caption{The segmentation process. 1. The candidate cut tokens are first inserted into the raw text via a simple rule (
newline in the example). 2. The initial text segments are fed into the segmenter, which outputs a binary probability distribution for each candidate cut token. 3. The segmenter can be applied recursively with different division thresholds to customize the segmentation granularity.  4. The segmentation for each candidate cut token is determined by the corresponding consecutive cut token.}
\label{fig.segmentation}
\vspace{-0.4cm}
\end{wrapfigure}

To enable general automatic segmentation, we adopt a data-driven approach and first construct a semantic segmentation dataset, called \textit{SemanticSeg}. Then we design the segmenter and its processing approach, and use the built dataset to train it.
\vspace{-0.4cm}

\subsection{SemanticSeg}
We diversify the source and length ($l \in [2k, 32k]$) of the dataset to facilitate the generalization of the segmenter. For each input text, we follow step 1 in Fig. \ref{fig.segmentation} to insert candidate cut tokens and use Gemini-2.5-Pro to determine the final segmentation results. SemanticSeg contains around 16 categories of segmentation data, with each category containing at least around $2k$ instances. The varying cut rates across categories can also help the segmenter learn distinct segmentation patterns. More details of the dataset can be found in Appendix \ref{appendix.semanticseg}.

\subsection{Segmentation}

We construct the segmenter with a pre-trained language model backbone (Qwen3-4B-Instruct-2507) by adding a classification head consisting of two linear layers and an intermediate ReLU activation layer. The segmentation process is presented in Fig. \ref{fig.segmentation}. For an input text $T$, a set of candidate cut tokens $C\in\{C_0, C_1, \dots, C_n\}$ is first inserted into the input text via simple rules like the newline character.
This forms a series of initial text segments $\{C_0, T_1, C_1, T_2, C_2, \dots, T_n, C_n\}$, where $\{T_1, T_2, \dots, T_n\}=T$. The segmenter then takes this series of text segments, along with the candidate cut tokens, as input and outputs a binary probability distribution for each candidate cut token. As the current candidate cut point cannot capture important segmentation information from its successive text segments (For example, in Fig. \ref{fig.segmentation}, "<cut 3>" cannot attend to its successive token "Method", which is a promising choice for final segmentation), we use the hidden vector from the next candidate to determine the segmentation of the current candidate. 

A very important factor of the segmentation is the granularity, which determines the final number of blocks (the parallel degree) and how much information each block contains. In the segmenter, two adjustable components could be used to control the segmentation granularity, which are the threshold value for the binary probability distribution and the recursion depth (the number of times the segmenter is applied recursively, with each level splitting existing blocks further using a pre-defined threshold) in the segmenter. The users can pair each level of recursion with a different threshold value. Generally, the deeper recursion level can pair with a greater or equal threshold value. During training, the threshold value is set to 0.5, but we recommend 0.2 \textasciitilde\,0.5 for the first level of recursion (An example is provided in Appendix \ref{appendix.segmentation-example}).

\section{Block Distillation}

Block fine-tuning \citep{block-attention} trains the model under block and full attention to preserve full-attention performance. This is time‑consuming and hard to scale. Therefore, we propose block distillation, which uses the original full-attention model to guide block-attention training. With block distillation, we can safely bypass the heavy updating scheme of block fine-tuning without degrading full-attention performance while improving block-attention performance (section \ref{sec.main-results}).

The block distillation employs three novel components to facilitate the training. They are \textit{block sink tokens} that are used to mitigate abnormal patterns in block attention, \textit{block dropout} that takes advantage of the non-last block training signal, and \textit{token weighting} applied to the token dimension to the cross-entropy loss.

\subsection{Block Sink Tokens}
The previous investigation \citep{attention-entropy} reveals that the attention patterns are extremely abnormal at the beginning of each block, leading to potential optimization instabilities. We characterize this challenge as \textit{lost in block head} (section \ref{sec.lost in block head}). To tackle this problem, we introduce a new token ($<|block\_start|>$) called block sink token. Following \citep{attention-sink}, we duplicate the block sink token four times at the beginning of each block. Hence, the final block-attention version input follows the format $\{bls*4, B_1, bls*4, B_2, \dots, bls*4, B_n\}$, where $bls$ means the block sink token $<|block\_start|>$ and $B_r, r\in[1,n]$ is the blocks partitioned by the segmenter. In practice, we set the dropout rate to around 0.6 for all the training.

\begin{wrapfigure}{r}{0.4\textwidth}
\vspace{-0.4cm}
\centering
\includegraphics[width=0.4\textwidth]{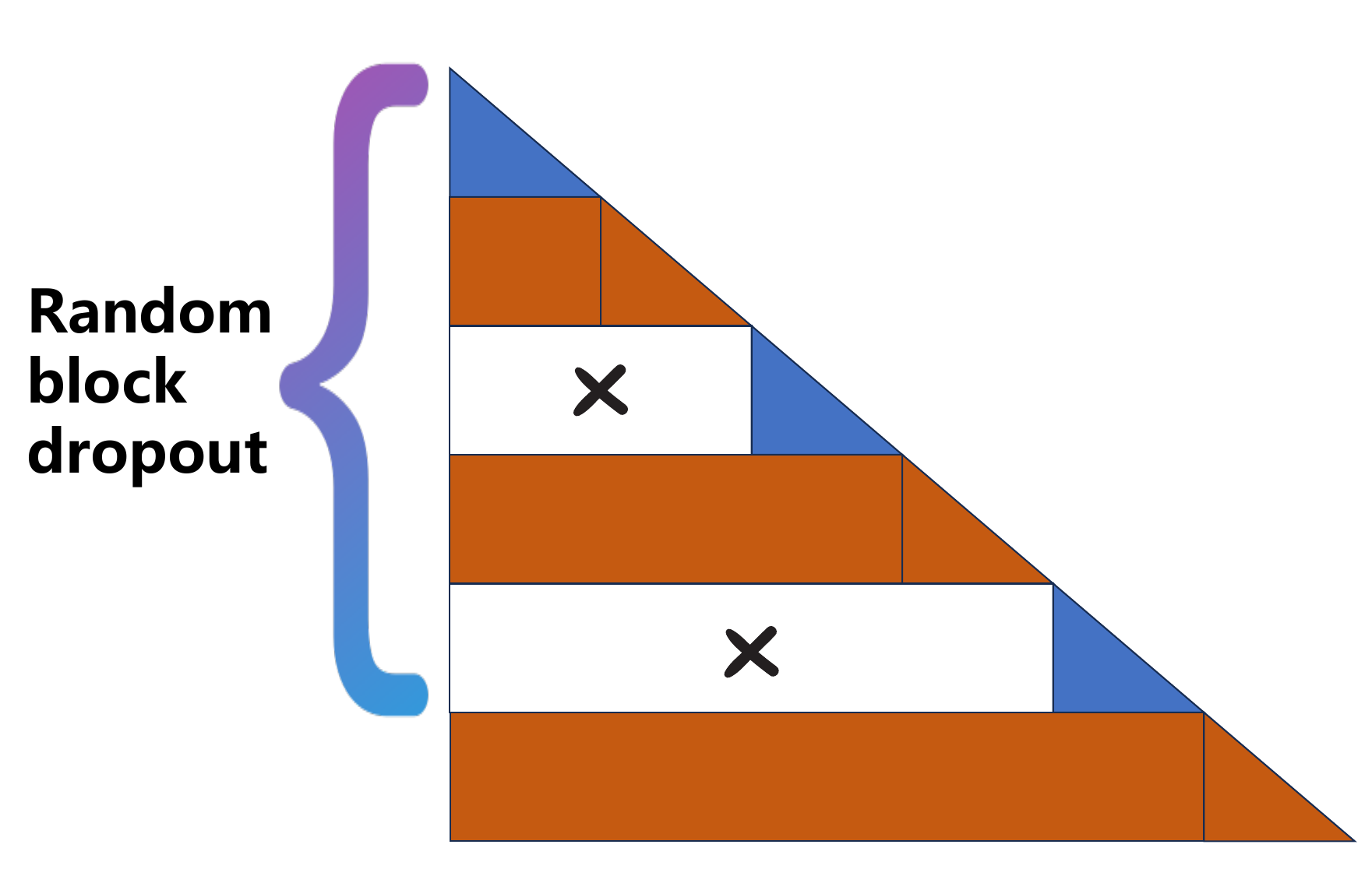}
\caption{The block dropout. A number of randomly selected blocks are forced to attend only the content within the block itself. Note that the final block always follows the full-attention pattern.}
\label{fig.block-dropout}
\vspace{-0.4cm}
\end{wrapfigure}

\subsection{Block Dropout}
A fundamental requirement for block attention is the model's ability to accurately retrieve information from the KV caches of all the blocks. Existing fine-tuning methods \citep{block-attention} are highly inefficient because they only optimize the model using signals from the final block, essentially ignoring the vast majority of the input sequence. To address this signal sparsity problem, we introduce block dropout (Fig. \ref{fig.block-dropout}). This mechanism randomly selects a subset of context blocks for individual encoding (blue in Fig. \ref{fig.block-dropout}) and applies a KL divergence loss to all remaining non-corrupted blocks (orange). By doing so, we force the model to learn from a much larger proportion of the text. Formally, given an input sequence $x$, a frozen teacher model $\varphi$, a student model $\varphi_s$, and let $\mathfrak{R}(x)$ denote the set of tokens within corrupted blocks. The block dropout KL divergence is defined as:
\begin{gather}
\label{eq. block dropout kl}
KL_{x} = D_{KL}(p_\varphi(\{x_i | x_i \notin \mathfrak{R}(x)\}) \ || \ p_{\varphi_s}(\{x_i | x_i \notin \mathfrak{R}(x)\}))
\end{gather}

\subsection{Token Weighting}
Traditional cross-entropy loss applies equal weights to the token dimension. Such a mechanism potentially decreases the training effectiveness, as it contains no information on the different degrees of importance for each token. Thus, we introduce the token weights that are employed on the token dimension of the cross-entropy computation, similar to \citep{incomes}. Specifically, given a teacher model $\varphi$ whose weights are frozen, an input $x$, the token weights are computed as follows:
\begin{gather}
w_x = max(CE(\varphi_b(x)) - CE(\varphi(x)), 0) \times \alpha + \beta
\end{gather}
where $\varphi_b$ means a block-attention forward pass in the teacher model and $CE$ is the cross-entropy loss. The token weights assign greater weight to tokens that have a relatively large difference in loss between the block-attention and full-attention forward passes, and shrink the loss scale of those insensitive tokens ($CE(\varphi_b(x)) - CE(\varphi(x) \leq 0$) to $\beta$ (usually a value near 0.1). In this way, we can alleviate the noise in training and focus more on the learning of block-attention capability. We set $\alpha=0.2, \beta=0.1$ for Qwen series models and $\alpha=0.5, \beta=0.1$ for Llama series models in the later experiments.

\subsection{Training}
The final training loss is the combination of the previously introduced components. Specifically, 
\begin{gather}
loss_x = CE(\varphi_{bs}(x)) \times w_x + KL_x
\end{gather}
where $\varphi_{bs}$ represents a block-attention forward pass in the student model. More training details have been put into Appendix \ref{appendix.block-distillation-training}.

\section{Experiments}
In this section, we conduct comprehensive experiments to verify the key arguments of this paper from two perspectives. From the perspective of the segmentation: 1) The degree to which segmentation influences downstream performance, and 2) whether our segmenter offers a genuine improvement over other straightforward partition methods. From the perspectives of block distillation: 1) Whether block fine-tuning is enough for the general domain application, 2) whether the block distillation can help block-attention performance approach that of full attention in the general domain, 3) whether the block distillation affect the full attention performance, 4) what is the efficiency gain brought by the block attention in inference, and 5) the effectiveness of each component in block distillation.

\subsection{Experiment Settings}
\paragraph{Benchmarks}
We adopt two popular comprehensive benchmarks for evaluation, namely LongBench \citep{longbench} and LoCoMo \citep{locomo}, and follow the exact procedures defined in the original papers, including the metrics, prompts, etc.

\paragraph{Baselines}
Unlike \citep{block-attention}, which needs to reproduce the whole SFT procedure, we implement the block distillation directly on chat models, including Qwen3-4B-Instruct-2507, Llama-3.1-8B-instruct, Qwen3-8B, and Qwen3-14B. For fair comparison, we also include the model from previous work \citep{block-attention}. The details for the baselines are as follows:
\begin{itemize}
    \item \textbf{Original} - The untouched official model released on HuggingFace. This is the performance upper bound for the block attention model. Our objective is to make the general performance of the block attention model approximate that of this model as closely as possible.
    \item \textbf{Block-Dist} - The model trained via our block distillation framework using our segmented data.
    \item \textbf{Prompt Cache} - The Prompt Cache \citep{promptcache} baseline, which enables training-free reuse of the attention states across prompts.
    \item \textbf{Superposition} - The Superposition Prompting baseline \citep{Superposition-Prompting}. It allows LLMs to process the input documents in parallel paths, and prunes the irrelevant paths at the end.
    \item \textbf{Tulu3-SFT} - The original Llama-3.1-Tulu-3-8B-SFT model, which serves as the ceiling performance for the Tulu3 series of block attention baselines from \citep{block-attention}.
    \item \textbf{Tulu3-Block-FT} - The block attention model trained by block fine-tuning \citep{block-attention}, using the SFT dataset of Tulu3 and 20,000 samples of RAG data sampled from TriviaQA and 2WikiMultiHopQA. We include this model to visualize the gap between its block-attention performance and the full-attention performance from Tulu3-SFT in the general domain.
    \item \textbf{Tulu3-Block-FT-S} - Since the training data of Tulu3-Block-FT is partitioned by simple rules without the segmenter, we further train Tulu3-SFT using our data divided by the segmenter for fair comparison.
\end{itemize}
Unless otherwise specified, "- Full" indicates that the test is under full attention, and "- Block" means evaluation is using block attention.

\begin{table}[t]
\footnotesize
    \centering
    \setlength{\tabcolsep}{1.5mm}{
    \begin{adjustbox}{max width=\linewidth}
    \begin{tabular}{lccccccc}
    \toprule
    Method & Multi-doc QA & Single-doc QA & Summarization & Few shot & Synthetic & Code & Average \\ 
     \midrule
     \multicolumn{8}{c}{Qwen3-8B} \\
     \midrule
     Full & 40.52 & 47.51 & 24.01 & 35.59 & 66.99 & 01.66 & \colorbox{blue!30}{37.85} \\
     Random & 20.53 & 36.77 & 22.96 & 10.51 & 08.69 & 04.73 & 17.37 \\
     Average & 19.78 & 36.66 & 22.83 & 11.10 & 07.91 & 05.02 & 17.22 \\
     Punctuation & 06.14 & 11.62 & 21.55 & 08.72 & 02.85 & 06.25 & 09.52 \\
     Random candidate & 18.72 & 35.34 & 22.99 & 22.29 & 02.44 & 07.49 & 18.21 \\
     Average candidate & 17.24 & 33.11 & 22.83 & 21.86 & 02.32 & 07.01 & 17.40 \\
     Loss & 17.66 & 27.98 & 22.34 & 19.74 & 03.44 & 08.58 & 18.02 \\
     Entropy & 17.29 & 28.18 & 22.58 & 16.88 & 03.09 & 08.31 & 17.41 \\
     Segmenter & 17.88 & 36.92 & 22.81 & 24.05 & 03.65 & 04.61 & \colorbox{green!30}{18.32} \\
     \midrule
     \multicolumn{8}{c}{Tulu3-Block-FT} \\
     \midrule
     Tulu3-SFT - Full & 42.09 & 45.03 & 25.24 & 38.20 & 65.67 & 24.61 & \colorbox{blue!30}{40.40} \\
     Random & 36.96 & 40.49 & 15.30 & 27.13 & 19.57 & 29.89 & 30.38 \\
     Average & 36.05 & 40.52 & 24.75 & 28.95 & 15.35 & 27.66 & 28.88 \\
     Punctuation & 10.76 & 13.27 & 22.24 & 19.39 & 08.52 & 17.47 & 15.28 \\
     Random candidate & 36.84 & 40.40 & 25.17 & 45.00 & 03.12 & 30.74 & 30.21 \\
     Average candidate & 35.58 & 39.82 & 25.02 & 45.24 & 03.49 & 30.16 & 31.12 \\
     Loss & 32.66 & 34.94 & 24.67 & 36.46 & 05.41 & 28.49 & 28.01 \\
     Entropy & 30.31 & 33.76 & 24.59 & 35.87 & 3.83 & 32.40 & 27.35 \\
     Segmenter & 34.33 & 42.93 & 24.93 & 45.34 & 12.17 & 31.28 & \colorbox{green!30}{32.82} \\
     \midrule
    \bottomrule
    \end{tabular}
    \end{adjustbox}}
    \caption{Results for different segmentation methods on Longbench \citep{longbench}. For fair comparison, the parallel degree for all methods is aligned with that of the segmenter.}
    \label{tab:segmentation}
    \vspace{-0.4cm}
\end{table}

\begin{table}[t]
\footnotesize
    \centering
    \setlength{\tabcolsep}{1.5mm}{
    \begin{adjustbox}{max width=\linewidth}
    \begin{tabular}{lccccccc}
    \toprule
    Method & Multi-doc QA & Single-doc QA & Summarization & Few shot & Synthetic & Code & Average \\ 
     \midrule
    \multicolumn{8}{c}{Llama-3.1-Tulu-3-8B-SFT} \\
    \midrule
     Tulu3-SFT - Full & 42.09 & 45.03 & 25.24 & 38.20 & 65.67 & 24.61 & \colorbox{blue!30}{40.40} \\
     Tulu3-Block-FT - Block & 34.33 & 42.93 & 24.93 & 45.34 & 12.17 & 31.28 & 32.82 \\
     Tulu3-Block-FT-S - Block & 52.11 & 31.93 & 24.43 & 36.23 & 15.12 & 39.21 & 33.17 \\
     \midrule
    \bottomrule
    \end{tabular}
    \end{adjustbox}}
    \caption{Main results on Block-FT. " - Full" means the evaluation is performed using full attention, and "- Block" means the evaluation is performed under block attention.}
    \label{tab:Block-FT results}
    \vspace{-0.8cm}
\end{table}

\begin{table}[t]
\footnotesize
    \centering
    \setlength{\tabcolsep}{1.5mm}{
    \begin{adjustbox}{max width=\linewidth}
    \begin{tabular}{lccccccc}
    \toprule
    Method & Multi-doc QA & Single-doc QA & Summarization & Few shot & Synthetic & Code & Average \\ 
     \midrule
     \multicolumn{8}{c}{Qwen3-4B-Instruct-2507} \\
     \midrule
     Original - Full & 36.88 & 46.05 & 22.08 & 61.63 & 66.27 & 04.89 & \colorbox{blue!30}{39.63} \\
     Original - Prompt Cache & 28.36 & 31.89 & 17.89 & 49.23 & 48.49 & 04.83 & 30.12 \\
     Original - Superposition & 25.39 & 34.85 & 18.38 & 41.52 & 45.72 & 02.32 & 28.02 \\
     Block-Dist - Full & 38.74 & 47.14 & 22.33 & 59.22 & 67.66 & 12.70 & 41.29 \\
     Block-Dist - Block & 37.55 & 46.17 & 24.14 & 59.00 & 66.32 & 11.98 & \colorbox{green!30}{40.86} \\
     \midrule
     \multicolumn{8}{c}{Llama-3.1-8B-Instruct} \\
     \midrule
     Original - Full & 41.75 & 47.05 & 25.86 & 24.16 & 65.91 & 15.77 & \colorbox{green!30}{36.75} \\
     Original - Prompt Cache & 33.28 & 39.87 & 22.41 & 19.45 & 50.76 & 14.83 & 30.10 \\
     Original - Superposition & 31.95 & 37.81 & 25.75 & 21.45 & 46.15 & 13.87 & 29.50 \\
     Block-Dist - Full & 42.00 & 47.42 & 25.88 & 25.67 & 66.50 & 20.11 & \colorbox{blue!30}{37.93} \\
     Block-Dist - Block & 41.88 & 46.98 & 25.35 & 24.96 & 66.87 & 11.89 & 36.32 \\
     \midrule
     \multicolumn{8}{c}{Qwen3-8B} \\
     \midrule
     Original - Full & 40.52 & 47.51 & 24.01 & 35.59 & 66.99 & 01.66 & 36.05 \\
     Original - Prompt Cache & 28.89 & 38.86 & 23.77 & 20.76 & 45.27 & 05.67 & 27.20 \\
     Original - Superposition & 29.75 & 35.86 & 28.96 & 22.78 & 49.65 & 08.51 & 29.25 \\
     Block-Dist - Full & 42.75 & 47.11 & 25.75 & 38.97 & 67.33 & 11.33 & \colorbox{blue!30}{38.87} \\
     Block-Dist - Block & 41.77 & 46.38 & 25.17 & 36.76 & 67.00 & 09.95 & \colorbox{green!30}{37.83} \\
     \midrule
     \multicolumn{8}{c}{Qwen3-14B} \\
     \midrule
     Original - Full & 46.25 & 48.86 & 23.94 & 62.44 & 69.17 & 05.66 & \colorbox{green!30}{42.72} \\
     Original - Prompt Cache & 32.48 & 35.80 & 24.58 & 52.83 & 49.42 & 08.64 & 33.96 \\
     Original - Superposition & 35.88 & 32.85 & 22.77 & 48.54 & 45.23 & 10.41 & 32.61 \\
     Block-Dist - Full & 47.23 & 47.75 & 22.87 & 63.45 & 70.05 & 17.71 & \colorbox{blue!30}{44.84} \\
     Block-Dist - Block & 46.33 & 46.54 & 22.89 & 63.76 & 69.23 & 06.43 & 42.53 \\
     \midrule
    \bottomrule
    \end{tabular}
    \end{adjustbox}}
    \caption{Main results on LongBench \citep{longbench}.}
    \label{tab:main longbench results}
    \vspace{-0.8cm}
\end{table}

\begin{table}[t]
\footnotesize
    \centering
    \setlength{\tabcolsep}{1.5mm}{
    \begin{adjustbox}{max width=\linewidth}
    \begin{tabular}{lcccccc}
    \toprule
    Method & Multi hop & Single hop & Temporal & Open domain & Adversarial & Average \\ 
     \midrule
     \multicolumn{7}{c}{Qwen3-4B-Instruct-2507} \\
     \midrule
     Original - Full & 39.98 & 33.38 & 09.60 & 48.83 & 34.75 & \colorbox{blue!30}{33.31} \\
     Original - Prompt Cache & 19.78 & 20.24 & 05.76 & 31.57 & 19.64 & 19.40 \\
     Original - Superposition & 17.38 & 19.51 & 07.42 & 28.40 & 22.85 & 19.11 \\
     Block-Dist - Full & 38.86 & 34.09 & 09.93 & 49.34 & 32.34 & \colorbox{green!30}{32.91} \\
     Block-Dist - Block & 37.05 & 32.67 & 12.60 & 48.57 & 32.02 & 32.58 \\
     \midrule
     \multicolumn{7}{c}{Llama-3.1-8B-Instruct} \\
     \midrule
     Original - Full & 49.76 & 38.67 & 18.65 & 66.15 & 16.82 & \colorbox{blue!30}{38.01} \\
     Original - Prompt Cache & 32.61 & 25.75 & 10.75 & 50.65 & 15.37 & 27.03 \\
     Original - Superposition & 25.51 & 28.64 & 15.51 & 40.48 & 17.42 & 25.51 \\
     Block-Dist - Full & 50.25 & 36.57 & 17.82 & 67.72 & 17.23 & 37.92 \\
     Block-Dist - Block & 48.55 & 36.82 & 19.79 & 65.22 & 20.73 & \colorbox{green!30}{38.22} \\
     \midrule
     \multicolumn{7}{c}{Qwen3-8B} \\
     \midrule
     Original - Full & 40.44 & 32.43 & 18.28 & 58.30 & 23.32 & \colorbox{green!30}{34.55} \\
     Original - Prompt Cache & 29.71 & 25.62 & 15.37 & 48.27 & 17.51 & 27.30 \\
     Original - Superposition & 20.51 & 21.41 & 11.24 & 39.39 & 19.48 & 22.41 \\
     Block-Dist - Full & 39.08 & 34.03 & 19.87 & 56.28 & 23.99 & \colorbox{blue!30}{34.65} \\
     Block-Dist - Block & 38.76 & 32.71 & 16.26 & 57.38 & 22.49 & 33.52 \\
     \midrule
     \multicolumn{7}{c}{Qwen3-14B} \\
     \midrule
     Original - Full & 38.95 & 33.16 & 14.26 & 60.45 & 37.44 & \colorbox{green!30}{36.85} \\
     Original - Prompt Cache & 29.68 & 28.87 & 07.34 & 51.36 & 22.97 & 28.04 \\
     Original - Superposition & 27.62 & 20.63 & 04.39 & 42.59 & 19.41 & 22.93 \\
     Block-Dist - Full & 39.93 & 34.93 & 13.06 & 61.10 & 36.61 & \colorbox{blue!30}{37.13} \\
     Block-Dist - Block & 37.95 & 33.82 & 14.73 & 59.89 & 37.30 & 36.74 \\
     \midrule
    \bottomrule
    \end{tabular}
    \end{adjustbox}}
    \caption{Main results on LoCoMo \citep{locomo}.}
    \label{tab:main locomo results}
    \vspace{-0.4cm}
\end{table}

\subsection{Main Results}
\label{sec.main-results}

\paragraph{The impact of segmentation}
\label{sec.impact-segmentation}
In this section, we verify two points: How much impact does the segmentation have on the performance, and is the segmenter really better than other simple segmentation baselines? These two points are verified by comparing the segmenter with other segmentation methods (Table \ref{tab:segmentation}). Specifically, we include two sets of baselines, one of which is segmentation in \textit{heuristics}:
\begin{itemize}
    \item \textit{Random} - Segmentation in random with the set of candidate cut points to be the space.
    \item \textit{Average} - Segmentation in average with the set of candidate cut points to be the space.
    \item \textit{Punctuation} - Segmentation using the set of candidate cut points to be the sentence-ending punctuation.
    \item \textit{Random candidate} - Random segmentation with the set of candidate cut points to be in step 2 of Fig. \ref{fig.segmentation}.
    \item \textit{Average candidate} - Average segmentation with the set of candidate cut points to be in step 2 of Fig. \ref{fig.segmentation}.
\end{itemize}
Another set is segmentation in \textit{statistics}:
\begin{itemize}
    \item \textit{Loss} - The segmentation with the chunked topk cross-entropy loss value\footnote{We use the chunked topk to prevent cut points from clustering together}.
    \item \textit{Entropy} - The segmentation with the chunked topk token entropy value.
\end{itemize}
To exclude the impact of segmentation during training, we do not use models trained via block distillation. Instead, we apply these methods on two models, namely, Qwen3-8B - the original chat model, and Tulu3-Block-FT - the block attention model trained via block fine-tuning \citep{block-attention}. The parallel degree of all segmentation methods is aligned with that of the segmenter. 

The performance variance of different segmentation baselines from Tulu3-Block-FT is noticeable, demonstrating the impact of segmentation methods. Since Qwen3-8B is not trained on any segmented data, the performance variance of Qwen3-8B is generally lower than that of Tulu3-Block-FT. Although these two models do not use any training data partitioned by the segmenter, they both perform the best on average when the input is processed by the segmenter (in comparison with other segmentation baselines\footnote{Note that the random candidate and the average candidate generally follow the segmentation methods used by Tulu3-Block-FT.}). This significantly demonstrates the effectiveness of the segmenter.

\paragraph{Generalization failure of block FT}
\label{sec.weaknesses of block attention}
The previous work \citep{block-attention} only tests the block attention under the RAG scenario. Hence, how the block fine-tuning performs for the general domain remains an unverified point. Therefore, we first test the block attention model trained in the previous work \citep{block-attention} to find out whether it can achieve similar performance to the full attention model. The results are shown in Table \ref{tab:Block-FT results}. The Tulu3-Block-FT model has degraded block attention performance compared to that of the Tulu3-SFT - Full. There are two possibilities for this: one is the incompetence of the block fine-tuning, and the other is the difference in the training data that is segmented via simple rules. To eliminate the influence of the training data, we further train the "Tulu3-SFT" model using our training data partitioned by our segmenter, the model denoted as "Tulu3-Block-FT-S". The results show that the block-attention performance of this model ("Tulu3-Block-FT-S - Block") has a relatively noticeable gap compared to the full-attention performance of "Tulu3-SFT". Therefore, we can conclude that the source of the performance gap is the block fine-tuning, and it is not enough for the generalization of block attention.

\paragraph{Effectiveness of block distillation}
In this section, we verify the effectiveness of block distillation in both block attention and full attention. The results are shown in Table \ref{tab:main longbench results} and Table \ref{tab:main locomo results}. The Prompt Cache \citep{promptcache} and Superposition prompting \citep{promptcache} considerably degrade the model performance compared to vanilla full-attention. This aligns with the findings in \citep{block-attention} and section \ref{sec.impact-segmentation} that the model struggles to adapt the block attention pattern without training. For all models, both the block-attention and full-attention performance of block distillation achieves near-equivalent performance to that of full-attention from the original model. Therefore, the block distillation can help improve the block-attention capability of the model while preserving its full-attention ability.

\paragraph{Efficiency}
We measure both training and inference efficiency (Table \ref{tab:Efficiency results}). For training, block distillation requires $25,859.9ms$ per step, which is approximately 26\% faster than Block-FT ($34,941.1ms$), which demonstrates the efficiency of block distillation over block fine-tuning. For inference, we follow the input format $[context, user \, query]$ and measure the time-to-first-token (TTFT) for vanilla full-attention and block attention across context lengths from $8k$ to $64k$. The length of the user query is set to $200$. Block attention consistently achieves lower TTFT than vanilla full-attention, and the absolute gain grows with sequence length: the TTFT reduction increases from $57.9ms$ at $8k$ to $3,149.7ms$ at $64k$.

\subsection{Ablation study \& Analysis}

\begin{table}[t]
\footnotesize
    \centering
    \setlength{\tabcolsep}{1.5mm}{
    \begin{adjustbox}{max width=\linewidth}
    \begin{tabular}{lccccccc}
    \toprule
    Method & Multi-doc QA & Single-doc QA & Summarization & Few shot & Synthetic & Code & Average \\ 
     \midrule
     \multicolumn{8}{c}{Qwen3-4B-Instruct-2507} \\
     \midrule
     Block-Dist - Block & 37.55 & 46.17 & 24.14 & 59.00 & 66.32 & 11.98 & \colorbox{green!30}{40.86} \\
     - w/o Block sink tokens & 36.19 & 44.86 & 22.56 & 55.38 & 52.33 & 17.96 & 38.21 \\
     - w/o Block dropout & 36.21 & 45.22 & 23.71 & 45.35 & 43.23 & 13.23 & 34.49 \\
     - w/o KL loss & 38.26 & 38.22 & 21.56 & 39.67 & 40.87 & 14.25 & 32.13 \\
     - w/o Token weights & 42.87 & 39.87 & 23.78 & 59.29 & 45.50 & 20.99 & 38.71 \\
     \midrule
    \bottomrule
    \end{tabular}
    \end{adjustbox}}
    \caption{The ablation study results.}
    \label{tab:Alation study}
    \vspace{-0.8cm}
\end{table}

\paragraph{The lost in block head}
\label{sec.lost in block head}
Given such a segmented example from the Longbench \citep{longbench} synthetic task: 
\begin{quote}
    \vspace{-0.4cm}
    \item "Block 1 - Paragraph 1:\dots; Block 2 - Paragraph 2:\dots; Block 3 - Paragraph 3:\dots;\dots; Block n - The following is an abstract:\dots, Please enter the number of the paragraph that the abstract is from. The answer format must be like "Paragraph 1", "Paragraph 2", etc. The answer is: " 
\end{quote}
where the model is required to retrieve the information from the head (beginning) of the block. We find that the block attention model has serious trouble in dealing with this type of query. We name this phenomenon \textit{lost in block head}. We believe that the source of the problem is linked to the findings in a previous investigation \citep{attention-entropy}, which shows that the attention pattern is extremely abnormal at the head of the blocks. Interestingly, despite the enormously greater amount of training data used compared to this work, the block-FT model trained in \citep{block-attention} shows a considerable collapse in this synthetic task (Tulu3-Block-FT - Block in Table \ref{tab:main longbench results}), indicating that simple fine-tuning may not be enough.

\begin{wraptable}{l}{0.4\textwidth}
\footnotesize
    \centering
    \setlength{\tabcolsep}{5mm}{
    \begin{adjustbox}{max width=\linewidth}
    \begin{tabular}{lc}
    \toprule
    Method & Time (ms) \\ 
     \midrule
    \multicolumn{2}{c}{Training} \\
    \midrule
    Block-FT & 34,941.1 \\
    Block-Dist & 25,859.9 \\
    \midrule
    \multicolumn{2}{c}{Inference} \\
    \midrule
    TTFT-vanilla - $8k$ & 509.2 \\
    TTFT-block - $8k$ & 451.3 \\    
    TTFT-vanilla - $16k$ & 1,099.5 \\
    TTFT-block - $16k$ & 702.1 \\    
    TTFT-vanilla - $32k$ & 2,642.5 \\
    TTFT-block - $32k$ & 1,534.8 \\    
    TTFT-vanilla - $64k$ & 6,971.3 \\
    TTFT-block - $64k$ & 3,821.6 \\  
    \bottomrule
    \end{tabular}
    \end{adjustbox}}
    \caption{The efficiency measurement.}
    \label{tab:Efficiency results}
    \vspace{-0.6cm}
\end{wraptable}

\paragraph{Block sink tokens}
To tackle the lost in block head problem, we introduce a new special token "$<|block_start|>$", which is padded before each block's beginning to allieviate the abnormal attention pattern. We conduct an experiment to verify its effectiveness in Table \ref{tab:Alation study}. The performance shows a considerable decrease in the Synthetic task, which aligns with the discussion in section \ref{sec.lost in block head}. In addition, the few-shot and single-doc QA tasks also experience a noticeable drop, implying that the block sink token not only helps the understanding of the information in the block head but also the later actual block content.

\paragraph{Block dropout}
We verify the effectiveness of the block dropout and KL divergence on the Longbench benchmark \citep{longbench} (Table \ref{tab:Alation study}) via forcing computing the KL loss for the last block only. The results demonstrate a great decrease in the few-shot and synthetic tasks when the block dropout is absent during training, suggesting it helps alleviate the lost in block head problem. The KL loss component has also been verified by completely wiping it out. The results experience a tremendous drop in single-doc QA, few-shot, and synthetic tasks, suggesting its important role in helping learn block-attention patterns.

\paragraph{Token weights}
The effectiveness of the token weights is verified by using the usual mean reduction for the cross-entropy loss. Although adopting the cross-entropy without weights increases the Multi-doc QA performance, it contrarily degrades the performance in single-doc QA and synthetic tasks. 
\vspace{-0.4cm}

\section{Related work}
Block attention \citep{block-attention} partitions input sequences into independent blocks to enable efficient KV cache reuse and parallel prefilling, but its broader adoption is hindered by costly block fine‑tuning and limited generalization beyond RAG settings. Prompt Cache \citep{promptcache} explores training‑free modular attention reuse across prompts, CacheBlend \citep{cacheblend} reuses non-prefix KV cache with selective KV recompute, while Superposition prompting \citep{Superposition-Prompting} processes documents in parallel paths and prunes irrelevant ones. However, all approaches suffer from severe performance degradation when directly applied under block‑attention patterns without dedicated training. In contrast, our work introduces a learned semantic segmenter and an efficient block distillation framework that overcome these limitations, achieving block‑attention performance close to the full‑attention upper bound while preserving full‑attention capability.

\section{Conclusion}
In this work, we address two fundamental obstacles that prevent the broader adoption of block attention: the absence of a principled segmentation method and the inefficiency of existing block fine-tuning. To tackle the first, we construct SemanticSeg, a large-scale multi-domain segmentation dataset, and train a lightweight neural segmenter that partitions text into semantically coherent blocks with controllable granularity. To overcome the second, we propose Block Distillation, an efficient training framework that incorporates three novel components—block sink tokens, block dropout, and token-level loss weighting—to effectively transfer full-attention capability to the block-attention pattern. Extensive experiments on LongBench and LoCoMo across multiple model families demonstrate the effectiveness of the segmenter and Block Distillation.

\bibliography{custom}


\appendix

\section{Limitations}
There are several limitations of this work that we need to discuss.
\paragraph{Block dropout in pretraining}
So far, we have only explored employing the block dropout in the post-training phase. However, we believe it is possible to scale it to the pre-training phase to further improve the capability. At that time, the cross-entropy loss can also be applied to the non-corrupted block.
\paragraph{Thinking mode verification}
We do not verify the thinking mode function of the block attention model since the training data does not contain any thinking content. However, based on the investigation from previous work \citep{attention-entropy}, the thinking mode has high potential for being effective in block attention.
\paragraph{Model type \& size}
Due to resource limitations, we can only scale the model size to 14B. The effectiveness of block distillation to block attention under a larger model size requires further exploration. Additionally, the experiments are only conducted for dense models; the compatibility of block attention with other popular structures, like MoE, requires further discussion.
\paragraph{RL compatibility}
Reinforcement learning is a powerful tool for improving the model's reasoning and agentic capabilities nowadays. However, none of the existing works investigates the compatibility of block attention with RL. Therefore, we think this is a promising direction for further exploration. If they are proven compatible, then the cost of commercial models would be considerably reduced.
\paragraph{Agentic application}
This paper does not verify the effectiveness of block attention in agentic scenarios. However, a big advantage of block attention is the KV cache reuse across prompts. Considering the complicated context engineering operations for agents these days \citep{Pensieve-Paradigm}, if it can be applied to agentic applications such as multi-agent collaboration, it could save much waste on re-encoding, and the cost would be massively decreased. 
\paragraph{Integration with Sparse Attention}
Sparse attention has been widely studied these days \citep{nsa, deepseek-v4} to reduce the requirements for computation and KV cache storage in extremely long context scenarios. However, it inherits the KV cache reusing logic of vanilla attention, which means the reuse is confined to the prefix. Therefore, the integration of block attention and sparse attention would be a promising future direction to further reduce the computation and KV cache storage.
\paragraph{Integration with Loop Transformers}
The release of GPT6-Astra has brought the Loop transformers \citep{universal-transformers, test-time-latent-reasoning, loop-transformers-are-better, Reasoning-with-Latent-Thoughts, loopllm, loopformer} into attention. Compared to the Vallina transformer, the Loop transformer is likely to introduce two main changes: 1) The Massive increase in computation (especially for long context). 2) Possible massive decrease in output thinking tokens, which makes the computation in the prefilling stage the main barrier for scaling. We believe the nature of the block attention (KV cache reuse across prompts) is a promising cure to alleviate this barrier.

\section{More details about segmentation}

\subsection{SemanticSeg Dataset}
\label{appendix.semanticseg}
\begin{table}[t]
\footnotesize
    \centering
    \setlength{\tabcolsep}{1.5mm}{
    \begin{adjustbox}{max width=\linewidth}
    \begin{tabular}{lccc}
    \toprule
    Category & Source & Num & Cut rate \\ 
     \midrule
     Book chapter & Booksum \citep{booksum} & 3551 & 0.0851 \\
     Long instruction & LongAlphaca \citep{longalpaca} & 3895 & 0.0724 \\
     Short Paragraphs & MuSiQue \citep{musique} & 3254 & 0.9260 \\
     Chat history & LongMemEval \citep{longmemeval} & 3100 & 0.1315 \\
     Textbook chapter & TextbookChapters \citep{textbookchapters} & 1980 & 0.1031 \\
     Mathematical text & OpenWebMath \citep{openwebmath} & 1980 & 0.1259 \\
     ArXiv & SlimPajama \citep{slimpajama} & 1980 & 0.0268 \\
     Raw book & SlimPajama \citep{slimpajama} & 1980 & 0.0313 \\
     StackExchange QA & SlimPajama \citep{slimpajama} & 1980 & 0.0251 \\
     Educational web pages & FineWeb-Edu \citep{fineweb-edu} & 1980 & 0.1157 \\
     Wikipedia & SlimPajama \citep{slimpajama} & 1980 & 0.1015 \\
     Code - Comprehensive & The stack~\citep{the-stack} & 4821 & 0.2022 \\
     Code - Python & The stack~\citep{the-stack} & 1980 & 0.1190 \\
     Code - C & The stack~\citep{the-stack} & 1980 & 0.1227 \\
     Code - Java & The stack~\citep{the-stack} & 1980 & 0.1125 \\
     Code - Shell & The stack~\citep{the-stack} & 1980 & 0.1783 \\
     \midrule
    \bottomrule
    \end{tabular}
    \end{adjustbox}}
    \caption{The statistics for the SemanticSeg dataset. "Comprehensive" means all the existing code categories in The stack \citep{the-stack}. The cut rate is calculated as $\text{the number of authenticated cut tokens} \div \text{the number of candidate cut tokens}.$}
    \label{tab:SemanticSeg statistics}
\end{table}
The specific statistics and the resources for \textit{SemanticSeg} dataset are shown in Table \ref{tab:SemanticSeg statistics}. The prompt used for generating the segmentation data is as follows:
\begin{tcolorbox}[
    breakable,              
    verbatim,
    title={Prompt used by Gemini-2.5-Pro for \textit{SemanticSeg}},   
    colback=gray!5,      
    colframe=blue!75!white, 
    coltitle=black,        
    fonttitle=\bfseries,   
    center title  
]
You are a master editor specializing in text segmentation for parallel processing by Large Language Models. Your goal is to segment a given text into the largest possible, yet fully self-contained, semantic chunks. The text is pre-segmented with candidate markers `<cut 0>` to `<cut N>`.
\\
\\

Your task is to identify the correct boundaries by selecting a subset of these markers. A chunk is "self-contained" if a person (or an LLM) can understand it completely in isolation, without needing to read the preceding chunk.
\\
\\

**The Core Principle:**
The ideal cutting point is where the topic COMPLETELY changes, and the new chunk can be understood without any ambiguity. If a chunk starts with a sentence that makes you ask "wait, who/what/why are they talking about?", then the cut is wrong and the chunks must be merged.
\\
\\

**CRITICAL HEURISTICS FOR COHERENCE (Rules for what NOT to do):**
You must merge chunks avoid these common errors. A cut is BAD if:
\\

1.  **It splits a sentence.** The text after a "<cut *>" marker cannot be the grammatical continuation of a sentence from before the marker.
    Example: "...he demanded", "Do the Maquas dare..." -> BAD CUT. Merge the two chunks to form a complete sentence.
\\

2.  **It breaks direct anaphora (reference).** The new chunk cannot start with words or phrases that directly refer to the immediate preceding text.
    Examples: "I'll try those suggestions.", "Because of this...", "That's a great point.", "He/she then...", "As they..." -> BAD CUT. The reference ("those suggestions", "this", "That", "He/She", "They", etc.) is in the previous chunk. Merge the reference chunk with its immediately previous chunk.
\\

3.  **It separates a question from its answer, or a statement from its immediate response.** Dialogue is a continuous flow. A multi-turn exchange on a single, evolving topic should be in ONE chunk.
    Example: A user asks for coffee shops, the assistant lists them, the user asks about Wi-Fi at one of them. This entire thread about finding a coffee shop is ONE semantic unit and should be in the SAME chunk.
\\

4.  **It separates functionally dependent content.** Text that references an element (like a figure or table) must be in the same chunk as that element's description.
    Example: Text discussing "Figure 1" must be in the same chunk as the caption for "Figure 1".
\\

5.  **It separates the chunk content from its header.** The chunk content and its corresponding header should always stay in the same chunk.
    Example: "Chapter V\textbackslash n", "Mary did something bad...\textbackslash n" -> BAD CUT. The header ("Chapter", "Section", "Part", etc.) and its content should be in the SAME chunk. Merge the header chunk with its content chunk.
\\

6.  **It breaks the semantic clarity.** Each chunk should remain understandable to a downstream model or retrieval system.
    Examples: For Python, "def hello\_word(a:str)\textbackslash n", "   print("hello world!")" -> BAD CUT. The function head and its content should not be separated; "class myclass()\textbackslash n", "    def \_\_init\_\_(self, config):" -> BAD CUT. The class head and its content should not be separated. Merge them.
\\
\\

**Cutting steps you should refer:**
1.  Read the entire text to get a general sense of its structure (narrative, dialogue, academic paper, code, etc.).
2.  Iterate through each candidate marker from `<cut 1>` to `<cut N-1>`.
3.  For each marker, examine the two chunks immediately before and after it. Ask yourself whether this marker violates the heuristics above and whether it fulfills the core principle above.
4.  If the marker violates any of the above heuristics or does not fulfill the above core principle, identify this as a BAD cut point, move to the next marker and repeat step 3. Otherwise, identify this marker as a GOOD cut point.
5.  Based on the identified GOOD cut points, construct your final output.
\\
\\

**NOTE:**
(1). Usually, the final number of chunks should be greater than or equal to 5. 
(2). For the length of each chunk, it should contain no more than 350 candidate markers.
(3). Try your best to fulfill both (1) and (2). If you can't, you can compromise between (1) and (2) depend on the circumstances.
(4). Extremely large chunk is not allowed! Putting extremely large content into a chunk (For example, putting a whole paper into a chunk) to avoid BAD cuts is unacceptable!
\\
\\

**Output Format:**
Follow these rules exactly:
1.  Chunk boundaries must sit only on the exact strings `<cut *>`.
2.  Output **only** lines of the form and **nothing else**:
    `chunk <number>: <cut *> --- <cut **>`
    where `<cut *>` and `<cut **>` are the beginning and ending cutting point markers of the current chunk.
3.  Number chunks sequentially starting with 1.
4.  The beginning marker of the first chunk must be `<cut 0>` and the final marker of the final chunk must be `<cut N>`.
5.  Do not output any duplicated chunks.
\\
\\

Text to segment:
\end{tcolorbox}

\subsection{Segmentation Example}
\label{appendix.segmentation-example}
We provide a segmentation example about python code to facilitate the understanding. The example uses recursion depth 1 and threshold value 0.4.
\begin{tcolorbox}[
    breakable,              
    enhanced,    
    verbatim,
    title={An example processed by the segmenter},   
    colback=gray!5,      
    colframe=blue!75!white, 
    coltitle=black,        
    fonttitle=\bfseries,   
    center title  
]

chunk 1: \\
import numpy as np \\
from numpy.testing import assert\_array\_equal, assert\_array\_almost\_equal \\
import scipy.stats.distributions as distrs \\
from scipy.stats.kde import gaussian\_kde \\
from scipy.integrate import quad \\
import pytest
\\

chunk 2: \\
def augment\_grid(x, n\_inner\_points):
    test\_arr = [
        np.linspace(x[i], x[i + 1], n\_inner\_points + 1, endpoint=False)
        for i in np.arange(len(x) - 1)
    ]
    test\_arr.append([x[-1]])
    return np.concatenate(test\_arr)

def circle\_fun(x, low, high):
    x = np.array(x)
    center = 0.5 * (high + low)
    radius = 0.5 * (high - low)

    res = np.zeros\_like(x)

    center\_dist = np.abs(x - center)
    is\_in = center\_dist <= radius
    res[is\_in] = np.sqrt(radius ** 2 - center\_dist[is\_in] ** 2)

    return res
\\

chunk 3: \\
class TestCont:
"""Regression tests for `Cont` class"""

    def test\_init\_errors(self):
        def check\_one\_input(def\_args, var):
                with pytest.raises(TypeError, match=f"`{var}`.*numpy array"):
                    def\_args[var] = {"a": None}
                    Cont(**def\_args)
                with pytest.raises(TypeError, match=f"`{var}`.*float"):
                    def\_args[var] = ["a", "a"]
                    Cont(**def\_args)
                with pytest.raises(TypeError, match=f"`{var}`.*finite values"):
                    def\_args[var] = [0, np.nan]
                    Cont(**def\_args)
                with pytest.raises(TypeError, match=f"`{var}`.*finite values"):
                    def\_args[var] = [0, np.inf]
                    Cont(**def\_args)
                with pytest.raises(ValueError, match=f"`{var}`.*1d array"):
                    def\_args[var] = [[0, 1]]
                    Cont(**def\_args)

            check\_one\_input({"y": [1, 1]}, "x")
            check\_one\_input({"x": [0, 1]}, "y")

            with pytest.raises(ValueError, match="[Ll]engths.*match"):
                Cont([0, 1], [1, 1, 1])

            with pytest.raises(ValueError, match="two"):
                Cont([1], [1])

            with pytest.warns(UserWarning, match="`x`.*not sorted.*`x` and `y`"):
                rv = Cont([1, 0], [0, 2])
                rv\_ref = Cont([0, 1], [2, 0])
                \_test\_equal\_rand(rv, rv\_ref)

            with pytest.raises(ValueError, match="`y`.*negative"):
                Cont([0, 1], [1, -1])

            with pytest.raises(ValueError, match="`y`.*no positive"):
                Cont([0, 1], [0, 0])

chunk 4: \\
def test\_init(self):
        x\_ref = np.array([0, 1, 2])
        y\_ref = np.array([0, 1, 0])
        rv\_ref = Cont(x\_ref, y\_ref)
\\

chunk 5: \\
class TestFromRVAccuracy:
    """Accuracy of `Cont.from\_rv()`"""

    \# Output of `from\_rv()` should have CDF that differs from original CDF by
    \# no more than `thres`
    \@pytest.mark.slow
    \@pytest.mark.parametrize(
        "distr\_dict,thres",
        [
            (DISTRIBUTIONS\_COMMON, 1e-4),
            (DISTRIBUTIONS\_INF\_DENSITY, 1e-3),
            (DISTRIBUTIONS\_HEAVY\_TAILS, 5e-3),
        ],
    )
    def test\_cdf\_maxerror(self, distr\_dict, thres):
        test\_passed = {
            name: TestFromRVAccuracy.from\_rv\_cdf\_maxerror(distr) <= thres
            for name, distr in distr\_dict.items()
        }

        assert all(test\_passed.values())
\\

chunk 6: \\
class TestFromSampleAccuracy:
    """Accuracy of `Cont.from\_sample()`"""

    \# Output of `from\_sample()` should differ from original density estimate by
    \# no more than `thres` (with default density estimator)
    \@pytest.mark.slow
    \@pytest.mark.parametrize(
        "distr\_dict,thres",
        [
            (DISTRIBUTIONS\_COMMON, 1e-4),
            (DISTRIBUTIONS\_INF\_DENSITY, 1.5e-4),
            (DISTRIBUTIONS\_HEAVY\_TAILS, 1e-4),
        ],
    )
    def test\_close\_cdf(self, distr\_dict, thres):
        rng = np.random.default\_rng(101)
        test\_passed = {
            name: TestFromSampleAccuracy.simulated\_cdf\_error(distr, rng) <= thres
            for name, distr in distr\_dict.items()
        }
\end{tcolorbox}
The segmenter successfully recognizes the boundaries of the classes (chunk 3-6) included in the python code and it also identify those non-class boundaries (chunk 1-2), demonstrating its generalization capability.

\section{Training details}

\subsection{Segmenter}

We use all categories from the \textit{SemanticSeg} for the training of the segmenter. Note that for the code category, we only use the comprehensive subset. The segmenter structure is an autoregressive model backbone with a new cut head. The cut head consists of two linear layers and an intermediate ReLU activation layer. We use the learning rate $2e^{-5} - 2e^{-6}$ with a cosine decay strategy.

\subsection{Block distillation}
\label{appendix.block-distillation-training}
We use the flex attention \citep{flex-attention} framework and liger-kernel \citep{liger-kernel} to implement the block attention during training. For all models, we adopt a learning rate $2e^{-6} - 2e^{-7}$ and a cosine decay strategy.

For the training dataset, we adopt HotpotQA \citep{hotpotqa} and a subset from ChatQA2 \citep{chatqa2}. We use the trained segmenter to divide the samples from the subset of ChatQA2 (called ChatQA2Seg) for the training dataset. Overall, the number of training data is around $180k$, and the length is less than $32k$.

\section{Application scenario analysis}
\label{appendix.application-analysis}
\subsection{Coding agent}
Consider an LLM-powered coding assistant managing a large repository. A developer first asks Query A about python1.py and python2.py, and later asks Query B about python3.py and python4.py. Both queries share no overlapping files except the project’s global config.

Under full attention, the KV cache is context-dependent: it is tied to the exact sequence of documents in a specific prompt. If the assistant encodes the retrieved files for Query A, the cached KV states are affected by the specific file order and Query A’s tokens. To serve Query B with a different file subset or order, the cache cannot be partially reused; the entire prompt, including potentially many unchanged files, must be re-encoded. This leads to significant and wasteful recomputation.

Block attention decouples this dependency by treating each document as an independently encoded block. The KV states of python1.py, python2.py, python3.py, python4.py, etc., are stored separately. When Query B arrives, only the required blocks are fetched and composed with the new query block — no re-encoding of unchanged documents is needed. This modular, prompt-level KV cache reuse is the fundamental efficiency gain of block attention: it replaces rigid, monolithic caching with flexible, composable caching, dramatically reducing redundant prefilling in dynamic, long-context scenarios typical of coding agents and multi-document workflows.

Assume the input obeys the following format $[system \, prompt, retrieved \, code \, files, user \, query]$ and the length of the system prompt is around $15k$, the repositories contains around $30$ files with each file has around $10k$ length, the user query length is $200$, and we retrieve $10$ files for every user query. For vanilla prefix cache, only the system prompt can be reused, hence the cache hit rate $15k \div (15k+100k+200) \approx 0.13$. However, for block attention, since the code files are all encoded and cache separately, they can be reused directly when retrieved. The cache hit rate $(15k+100k) \div (15k+100k+200) \approx 0.99$. Given that the price with cache hit could be $10$ times lower than that of the missed hit, this can save tremendous computation time and cost.

\subsection{Multi-turn agentic workflows}
Consider an LLM‑based research agent tasked with "Investigate recent advances in mechanistic interpretability and summarize key findings." The agent operates in a multi‑turn ReAct‑style loop \citep{react}: at each step, it decides which tool to use (search, browse, or code execution), processes the results, and plans the next action. The specific turns are as follows:
\begin{quote}
    \textbf{Turn 1}: Search for "mechanistic interpretability survey 2025" and retrieve Paper A, Paper B, and Paper C. \\
    \textbf{Turn 2}: Browse Paper A and extract the main research landscape. \\
    \textbf{Turn 3}: Browse Paper B and understand the evidence and limitations. \\
    \textbf{Turn 4}: Browse Paper C and record its experimental design and conclusions. \\
    \textbf{Turn 5}: Revisit Paper A to reuse its taxonomy for structuring the final report. \\
    \textbf{Turn 6}: Revisit Paper B to compare its evidence with the results in Paper C. \\
    \textbf{Turn 7}: Extract key sentences from A, B, and C to build a comparison table. \\
    \textbf{Turn 8}: Compose the final report with specific citations to all three papers. 
\end{quote}
Under full attention, each new turn concatenates the growing history with newly retrieved documents and the agent's next action. Even if papers A, B, and C remain identical across turns, their KV states are embedded in a monolithic context that changes with every search result and reasoning step. Their cached states from previous turns are rarely reusable because the surrounding prompt context and document ordering have shifted, forcing repeated re‑encoding of the same documents.

With block attention, each permanent element (e.g., the system prompt, Paper A, Paper B, Paper C, etc) is encoded as an independent block and cached once. Each agent turn only requires encoding the new query, any new search results, and the fresh reasoning step, while the static document blocks are fetched directly from the cache and combined in a modular way. Over many turns and many documents, this eliminates repeated prefilling of unchanged content, yielding compounding efficiency savings and substantially lowering latency and compute cost in long‑running agentic workflows.

Suppose all paper A, B and C has $20k$ in length. For vanilla prefix cache, the cached states from previous turns can only be reused as a whole. However, for block attention, since the paper A, B and C are encoded independently, their states can be safely reused once they are prefilled in turn 2-4. Therefore, the number of cache hit tokens from turn 5-7 would have a increase of $20k \times 6 = 120k$.

\section{Societal impacts}
By boosting long-context efficiency and KV cache reuse, our work lowers compute cost and energy use, making advanced AI more accessible and sustainable for coding assistants, multi-document analysis, and agentic applications. However, cheaper inference may lower barriers for misuse (e.g., disinformation) and accelerate deployment without labor safeguards. Efficiency benefits may not transfer equally across architectures or languages, risking a widening gap between well-supported and underrepresented settings. Responsible deployment and monitoring are encouraged.



\end{document}